\documentclass[11pt,a4paper]{article}
\usepackage{authblk}
\usepackage[hyperref]{acl2021}
\usepackage{times}
\usepackage{latexsym}

\usepackage{enumitem}
\usepackage{soul}
\usepackage{url}
\usepackage{graphicx}
\usepackage{amsmath}
\usepackage{multirow}
\usepackage{amsthm}
\usepackage{booktabs}
\usepackage{algpseudocode}
\usepackage{algorithmicx}
\usepackage{algorithm}
\usepackage{IEEEtrantools}
\usepackage{amssymb, amscd, amsfonts}
\usepackage{subfigure}
\usepackage{arydshln}
\usepackage{makecell}
\usepackage[title]{appendix}

\newcommand{\tabincell}[2]{\begin{tabular}{@{}#1@{}}#2\end{tabular}} 

\makeatletter
\def\adl@drawiv#1#2#3{%
        \hskip.5\tabcolsep
        \xleaders#3{#2.5\@tempdimb #1{1}#2.5\@tempdimb}%
                #2\z@ plus1fil minus1fil\relax
        \hskip.5\tabcolsep}
\newcommand{\cdashlinelr}[1]{%
  \noalign{\vskip\aboverulesep
           \global\let\@dashdrawstore\adl@draw
           \global\let\adl@draw\adl@drawiv}
  \cdashline{#1}
  \noalign{\global\let\adl@draw\@dashdrawstore
           \vskip\belowrulesep}}
\makeatother

\newcommand\blfootnote[1]{%
  \begingroup
  \renewcommand\thefootnote{}\footnote{#1}%
  \addtocounter{footnote}{-1}%
  \endgroup
}

\usepackage{microtype}

\aclfinalcopy 
\def\aclpaperid{2436} 

\newcommand\mymodel{\textsc{UniRE}}

\title{\mymodel: A Unified Label Space for Entity Relation Extraction}

\author[1, 2]{\bf Yijun Wang$^*$}
\author[4]{\bf Changzhi Sun$^*$}
\author[3]{\bf Yuanbin Wu}
\author[4]{\bf Hao Zhou}
\author[4]{\bf Lei Li}
\author[1, 2]{\bf Junchi Yan$^\dagger$}
\affil[1]{Department of Computer Science and Engineering, Shanghai Jiao Tong University}
\affil[2]{MoE  Key  Lab  of  Artificial  Intelligence,  AI Institute, Shanghai Jiao Tong University}
\affil[3]{School of Computer Science and Technology, East China Normal University}
\affil[4]{ByteDance AI Lab}
\affil[  ]{\tt \{yjwang.cs, yanjunchi\}@sjtu.edu.cn ybwu@cs.ecnu.edu.cn}
\affil[  ]{\tt \{sunchangzhi, zhouhao.nlp, lileilab\}@bytedance.com}

\begin{document}
\maketitle
\begin{abstract}
        Many joint entity relation extraction models
        setup two separated label spaces for the two sub-tasks
        (i.e., entity detection and relation classification).
        We argue that this setting may hinder the information interaction
        between entities and relations.
        In this work, we propose to eliminate the different treatment
        on the two sub-tasks' label spaces.
        The input of our model is a table containing all word pairs from a sentence.
        Entities and relations are represented by squares and rectangles in the table.
        We apply a unified classifier to predict each cell's label,
        which unifies the learning of two sub-tasks.
        For testing,
        an effective (yet fast) approximate decoder is proposed
        for finding squares and rectangles from tables.
        Experiments on three benchmarks (ACE04, ACE05, SciERC) show
        that, using only half the number of parameters,
        our model achieves competitive accuracy with the best extractor,
        and is faster.

\end{abstract}

\blfootnote{$^*$Equal contribution.}
\blfootnote{$^\dagger$Corresponding Author.}

\section{Introduction}

\begin{figure}[tb!]
    \begin{center}
        \includegraphics[width=3.0in]{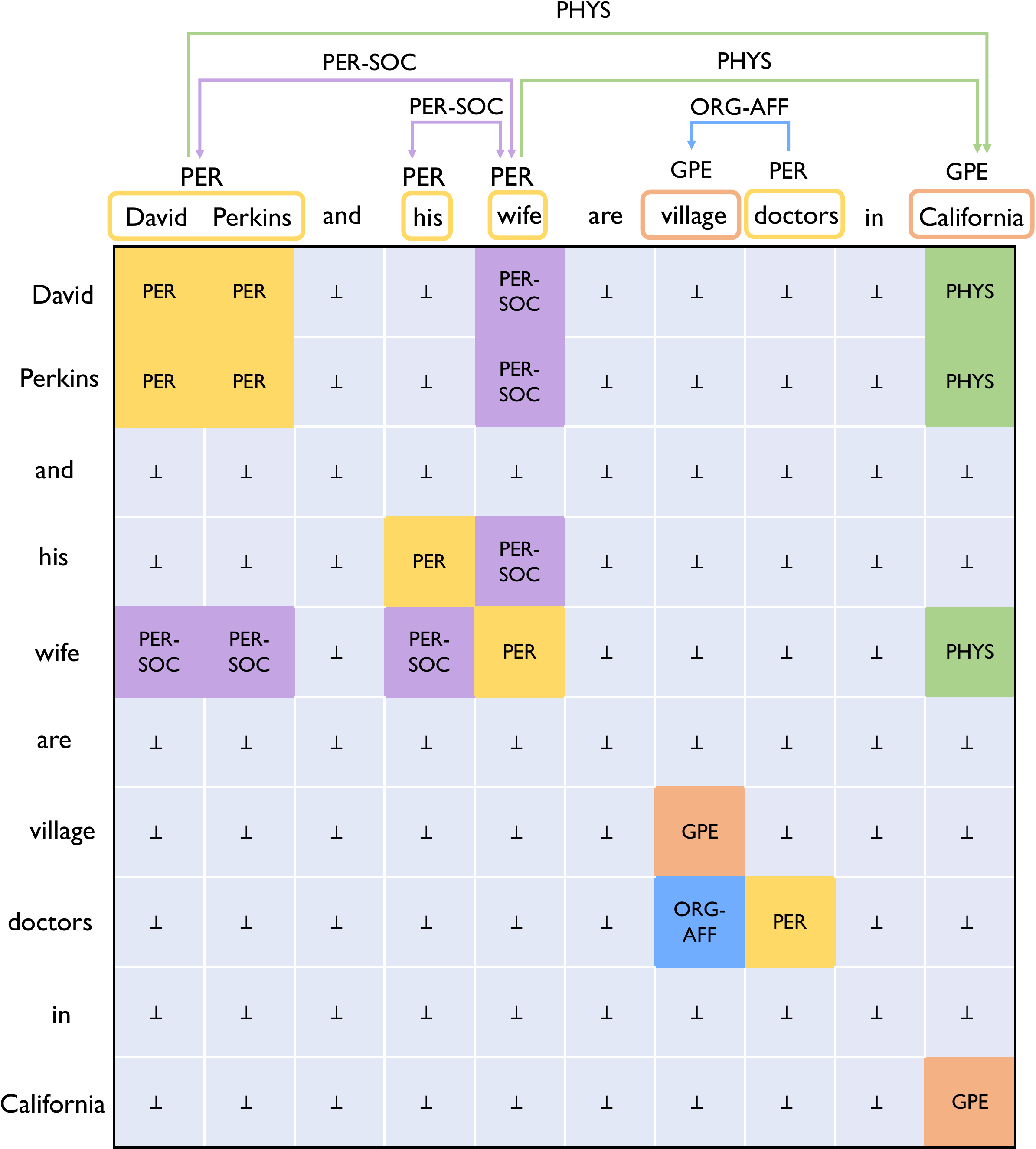}
    \end{center}
    \caption{Example of a table for joint entity relation extraction. 
    Each cell corresponds to a word pair. 
    Entities are squares on diagonal, relations are rectangles off diagonal.
    Note that \texttt{PER-SOC} is a undirected (symmetrical) relation type, 
    while \texttt{PHYS} and \texttt{ORG-AFF} are directed (asymmetrical) relation types.
    The table exactly expresses overlapped relations, e.g., the person entity ``David Perkins'' participates in two relations, (``David Perkins'', ``wife'', \texttt{PER-SOC}) and (``David Perkins'', ``California'', \texttt{PHYS}).
    For every cell, a \textbf{same} biaffine model predicts its label.
    The joint decoder is set to find the best squares and rectangles.
    }
    \label{fig:task}
\end{figure}

Extracting structured information from plain texts
is a long-lasting research topic in NLP.
Typically, it aims to recognize
specific entities and relations
for profiling the semantic of sentences.
An example is shown in \autoref{fig:task},
where a person entity ``David Perkins''
and a geography entity  ``California''
have a physical location relation \texttt{PHYS}.


Methods for detecting entities and relations
can be categorized into pipeline models or joint models.
In the pipeline setting,
entity models and relation models are independent
with disentangled feature spaces and output label spaces.
In the joint setting, on the other hand,
some parameter sharing of feature spaces
\citep{miwa-bansal-2016-end,katiyar-cardie-2017-going}
or decoding interactions
\citep{yang2013joint,sun-etal-2019-joint}
are imposed to explore the common structure of the two tasks.
It was believed that joint models
could be better since they can alleviate error propagations
among sub-models, have more compact parameter sets,
and uniformly encode prior knowledge (e.g., constraints) on both tasks.

However, \citet{zhong2020frustratingly} recently show
that with the help of modern pre-training tools (e.g., BERT),
separating the entity and relation model
(with independent encoders and pipeline decoding)
could surpass existing joint models.
They argue that,
since the output label spaces of entity and relation models are different,
comparing with shared encoders,
separate encoders could better capture
distinct contextual information,
avoid potential conflicts among them,
and help decoders making a more accurate prediction,
that is, \emph{separate label spaces deserve separate encoders}.

In this paper, we pursue a better joint model for
entity relation extraction.
After revisiting existing methods,
we find that though entity models and relation models share encoders,
usually their label spaces are still separate (even in models with joint decoders).
Therefore, parallel to \cite{zhong2020frustratingly},
we would ask whether
\emph{joint encoders (decoders) deserve joint label spaces}?

The challenge of developing a unified entity-relation label space
is that the two sub-tasks
are usually formulated into different learning problems
(e.g., entity detection as sequence labeling, relation classification
as multi-class classification),
and their labels are placed on different things
(e.g., words v.s. words pairs).
One prior attempt \cite{zheng2017joint} is to handle
both sub-tasks with one sequence labeling model.
A compound label set was devised to encode both entities and relations.
However, the model's expressiveness is sacrificed:
it can detect neither overlapping relations
(i.e., entities participating in multiple relation)
nor isolated entities (i.e., entities not appearing in any relation).

Our key idea of defining a new unified label space is that,
if we think
\citet{zheng2017joint}'s solution is to
perform relation classification during
entity labeling,
we could also consider the reverse direction by
seeing entity detection as a special case of relation classification.
Our new input space is a two-dimensional table with
each entry corresponding to a word pair in sentences (\autoref{fig:task}).
The joint model assign labels to each cell from \emph{a unified label space}
(union of entity type set and relation type set).
Graphically, entities are squares on the diagonal,
and relations are rectangles off the diagonal.
This formulation retains full model expressiveness regarding existing
entity-relation extraction scenarios (e.g., overlapped relations,
directed relations, undirected relations).
It is also different from the current table filling
settings for entity relation extraction
\citep{miwa2014modeling,gupta-etal-2016-table,zhang-etal-2017-end,wang-lu-2020-two},
which still have separate label space for entities and relations,
and treat on/off-diagonal entries differently.

Based on the tabular formulation,
our joint entity relation extractor performs two actions,
\emph{filling} and \emph{decoding}.
First, filling the table is to predict each word pair's label,
which is similar to arc prediction task in dependency parsing.
We adopt the biaffine attention mechanism \citep{dozat2016deep}
to learn interactions between word pairs.
We also impose two structural constraints on the table
through structural regularizations.
Next, given the table filling with label logits,
we devise an approximate joint decoding algorithm
to output the final extracted entities and relations.
Basically, it efficiently finds split points in the table
to identify squares and rectangles
(which is also different with existing table filling
models which still apply certain sequential decoding
and fill tables incrementally).

Experimental results on three benchmarks (ACE04, ACE05, SciERC) show that
the proposed joint method achieves competitive performances
comparing with the
current state-of-the-art extractors \cite{zhong2020frustratingly}:
it is better on ACE04 and SciERC, and competitive on ACE05.\footnote{Source code and models are available at \url{https://github.com/Receiling/UniRE}.}
Meanwhile, our new joint model is fast on decoding ($10$x faster than
the exact pipeline implementation, and comparable to an
approximate pipeline, which attains lower performance).
It also has a more compact parameter set:
the shared encoder uses only half the number of parameters comparing with
the separate encoder \cite{zhong2020frustratingly}.

\section{Task Definition}
Given an input sentence $s = x_1, x_2, \dots, x_{|s|}$ ($x_i$ is a word),
this task is to extract a set of entities $\mathcal{E}$ and a set of relations $\mathcal{R}$.
An entity $e$ is a span ($e.\texttt{span}$)
with a pre-defined type $e.\texttt{type} \in \mathcal{Y}_e$
(e.g., \texttt{PER}, \texttt{GPE}).
The span is a continuous sequence of words.
A relation $r$ is a triplet $(e_1, e_2, l)$,
where $e_1, e_2$ are two entities and $l \in \mathcal{Y}_r$ is a pre-defined relation type
describing the semantic relation among two entities (e.g., the \texttt{PHYS} relation between \texttt{PER} and \texttt{GPE} mentioned before).
Here $\mathcal{Y}_e, \mathcal{Y}_r$ denote the set of possible entity types and relation types respectively.

We formulate the joint entity relation extraction as
a table filling task (multi-class classification between each word pair in sentence $s$), as shown in \autoref{fig:task}.
For the sentence $s$,
we maintain a table $T^{|s| \times |s|}$.
For each cell $(i, j)$ in table $T$, we assign a label $y_{i, j} \in \mathcal{Y}$,
where $ \mathcal{Y} = \mathcal{Y}_e \cup \mathcal{Y}_r \cup \{\bot\} $ ( $\bot$ denotes no relation).
For each entity $e$, the label of corresponding cells $y_{i, j} (x_i \in e.\texttt{span}, x_j \in e.\texttt{span})$ should be filled in $e.\texttt{type}$.
For each relation $r = (e_1, e_2, l)$,
the label of corresponding cells $y_{i, j} (x_i \in e_1.\texttt{span}, x_j \in e_2.\texttt{span})$
should be filled in $l$.\footnote{Assuming no overlapping entities in one sentence.}
While others should be filled in $\bot$.
In the test phase, decoding entities and relations becomes a rectangle finding problem.
Note that solving this problem is not trivial,
and we propose a simple but effective joint decoding algorithm to tackle this challenge.

\begin{figure*}[t]
    \begin{center}
        \includegraphics[width=5in]{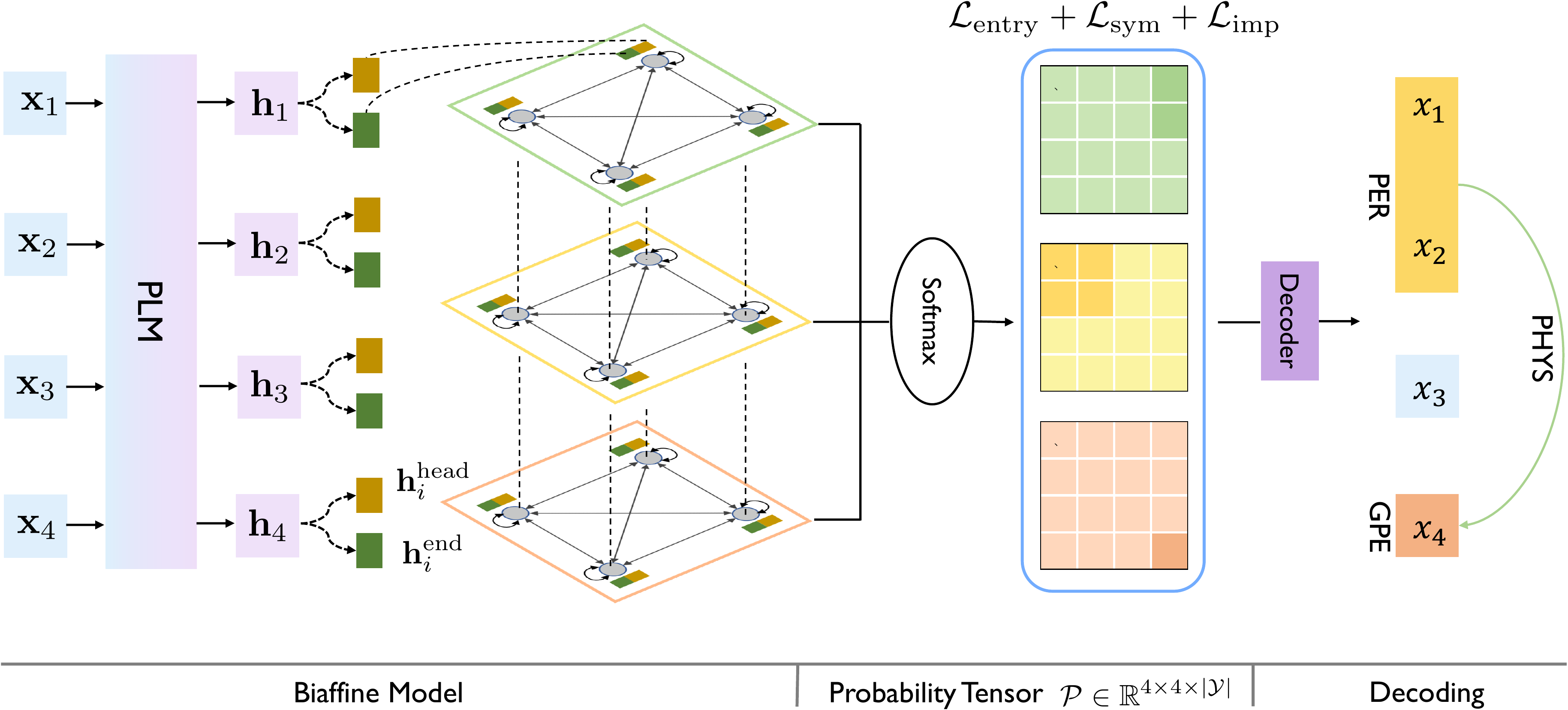}
    \end{center}
    \caption{Overview of our model architecture. One main objective ($\mathcal{L}_{\mathrm{entry}}$) and two additional objectives ($\mathcal{L}_{\mathrm{sym}},\mathcal{L}_{\mathrm{imp}}$) are imposed on probability tensor $\mathcal{P}$ and optimized jointly.}
    \label{fig:model}
\end{figure*}

\section{Approach}
In this section,
we first introduce our biaffine model for table filling task
based on pre-trained language models (Section \ref{biaffine-model}).
Then we detail the main objective function of the table filling task (Section \ref{table-filling})
and some constraints which are imposed on the table in training stage (Section \ref{constraints}).
Finally we present the joint decoding algorithm to extract entities and relations (Section \ref{joint-decoding}).
\autoref{fig:model} shows an overview of our model architecture.\footnote{We only show three labels of $\mathcal{Y}$ in \autoref{fig:model} for simplicity and clarity.}

\subsection{Biaffine Model} \label{biaffine-model}
Given an input sentence $s$,
to obtain the contextual representation $\mathbf{h}_i$ for each word,
we use a pre-trained language model (PLM) as our sentence encoder (e.g., BERT).
The output of the encoder is
\begin{IEEEeqnarray*}{c}
    \{\mathbf{h}_1, \dots, \mathbf{h}_{|s|}\} = \mathtt{PLM}(\{\mathbf{x}_1, \dots, \mathbf{x}_{|s|}\}),
\end{IEEEeqnarray*}
where $\mathbf{x}_i$ is the input representation of each word $x_i$.
Taking BERT as an example,
$\mathbf{x}_i$ sums the corresponding token, segment and position embeddings.
To capture long-range dependencies,
we also employ cross-sentence context following \citep{zhong2020frustratingly},
which extends the sentence to a fixed window size $W$
($W = 200$ in our default settings).

To better encode direction information of words in table $T$,
we use the deep biaffine attention mechanism \citep{dozat2016deep},
which achieves impressive results in the dependency parsing task.
Specifically,
we employ two dimension-reducing MLPs (multi-layer perceptron),
i.e., a head MLP and a tail MLP,
on each $\mathbf{h}_i$ as
\begin{IEEEeqnarray*}{c}
    \mathbf{h}_i^\mathrm{head} = \mathtt{MLP}_\mathrm{head}(\mathbf{h}_i), ~ ~
    \mathbf{h}_i^\mathrm{tail} = \mathtt{MLP}_\mathrm{tail}(\mathbf{h}_i),
\end{IEEEeqnarray*}
where $\mathbf{h}_i^\mathrm{head} \in \mathbb{R}^{d}$ and $\mathbf{h}_i^\mathrm{tail} \in \mathbb{R}^{d}$ are projection representations,
allowing the model to identify the head or tail role of each word.
Next, we calculate the scoring vector $\mathbf{g}_{i,j} \in \mathbb{R}^{|\mathcal{Y}|}$ of each word pair with biaffine model,
\begin{IEEEeqnarray*}{c}
    \mathbf{g}_{i,j} = \mathtt{Biaff}(\mathbf{h}_i^\mathrm{head}, \mathbf{h}_j^\mathrm{tail}), \\
    \mathtt{Biaff}(\mathbf{h}_1, \mathbf{h_2}) =
    \mathbf{h}_1^T \mathbf{U}_1 \mathbf{h}_2 + \mathbf{U}_2 (\mathbf{h}_1 \oplus \mathbf{h}_2) + \mathbf{b},
\end{IEEEeqnarray*}
where $\mathbf{U}_1 \in \mathbb{R}^{|\mathcal{Y}| \times d \times d}$ and $\mathbf{U}_2 \in \mathbb{R}^{|\mathcal{Y}| \times 2d}$ are weight parameters,
$\mathbf{b} \in \mathbb{R}^{|\mathcal{Y}|}$ is the bias,
$\oplus$ denotes concatenation.

\subsection{Table Filling} \label{table-filling}
After obtaining the scoring vector $\mathbf{g}_{i,j}$,
we feed $\mathbf{g}_{i, j}$ into the softmax function to predict corresponding label,
yielding a categorical probability distribution over the label space $\mathcal{Y}$ as
\begin{IEEEeqnarray*}{c}
    P(\mathbf{y}_{i, j} | s) = \mathtt{Softmax}(
    \mathtt{dropout}(\mathbf{g}_{i,j})).
    \label{eq:posterior}
\end{IEEEeqnarray*}
In our experiments,
we observe that applying dropout in $\mathbf{g}_{i, j}$,
similar to de-noising auto-encoding,
can further improve the performance.
\footnote{We set dropout rate $p = 0.2$ by default.}.
We refer this trick to \emph{logit dropout}
And the training objective is to minimize
\begin{IEEEeqnarray}{rrl}
    \mathcal{L}_\mathrm{entry} &=&- \frac{1}{|s|^2}\sum_{i = 1}^{|s|}{\sum_{j = 1}^{|s|}{\log P(\mathbf{y}_{i, j} = y_{i,j}|s)}},
    \label{loss-entry}
\end{IEEEeqnarray}
where the gold label $y_{i, j}$ can be read from annotations, as shown in \autoref{fig:task}.

\subsection{Constraints}
\label{constraints}
In fact, \autoref{loss-entry} is based on the assumption that each label is independent.
This assumption simplifies the training procedure, but ignores some structural constraints.
For example, entities and relations correspond to squares and rectangles in the table.
\autoref{loss-entry} does not encode this constraint explicitly.
To enhance our model,
we propose two intuitive constraints,
\emph{symmetry} and \emph{implication},
which are detailed in this section.
Here we introduce a new notation $\mathcal{P} \in \mathbb{R}^{|s| \times |s| \times |\mathcal{Y}|}$,
denoting the stack of $P(\mathbf{y}_{i, j}|s)$ for all word pairs in sentence $s$.\footnote{ $\mathcal{P}$ without \emph{logit dropout} mentioned in Section \ref{table-filling} to preserve learned structure.}

\paragraph{Symmetry}
We have several observations from the table in the tag level.
Firstly, the squares corresponding to entities must be symmetrical about the diagonal.
Secondly, for symmetrical relations,
the relation triples $(e_1, e_2, l)$ and  $(e_2, e_1, l)$ are equivalent,
thus the rectangles corresponding to two counterpart relation triples are also symmetrical about the diagonal.
As shown in \autoref{fig:task},
the rectangles corresponding to (``his'', ``wife'', \texttt{PER-SOC})
and (``wife'', ``his'', \texttt{PER-SOC}) are symmetrical about the diagonal.
We divide the set of labels $\mathcal{Y}$ into a symmetrical label set $\mathcal{Y}_\mathrm{sym}$ and an asymmetrical label set $\mathcal{Y}_\mathrm{asym}$.
The matrix $\mathcal{P}_{:, :, t}$ should be symmetrical about the diagonal for each label $t \in \mathcal{Y}_\mathrm{sym}$.
We formulate this tag-level constraint as symmetrical loss,
\begin{IEEEeqnarray*}{c}
    \mathcal{L}_\mathrm{sym} = \frac{1}{|s|^2}\sum_{i = 1}^{|s|}{\sum_{j = 1}^{|s|}{\sum_{t \in \mathcal{Y}_\mathrm{sym}}{|\mathcal{P}_{i, j, t} - \mathcal{P}_{j, i, t}|}}}.
\end{IEEEeqnarray*}
We list all $\mathcal{Y}_\mathrm{sym}$ in \autoref{tab:symmetrical-set} for our adopted datasets.

\begin{table}[tb!]
    \centering
    \footnotesize
    \resizebox{0.48\textwidth}{!}{
        \begin{tabular}{lcc}
            \toprule
            \multirow{2}{*}{Dataset} & \multicolumn{2}{c}{$\mathcal{Y}_\mathrm{sym}$ }                    \\
            ~                        & Ent                                                          & Rel \\
            \midrule
            \tabincell{c}{ACE04/                                                                          \\ ACE05}                    & \tabincell{c}{\texttt{PER},\texttt{ORG},\texttt{LOC},                     \\ \texttt{FAC},\texttt{WEA},\texttt{VEH},\texttt{GPE}}              & \texttt{PER-SOC} \\
            \midrule
            SciERC                   & \tabincell{c}{\texttt{Task},\texttt{Method},\texttt{Metric},       \\ \texttt{Material},\texttt{Generic}, \\\texttt{OtherScientificTerm}}
                                     & \tabincell{c}{\texttt{COMPAREP},                                   \\ \texttt{CONJUNCTION}  }                             \\
            \bottomrule
        \end{tabular}}
    \caption{Symmetrical label set $\mathcal{Y}_\mathrm{sym}$ for used datasets.}
    \label{tab:symmetrical-set}
\end{table}

\paragraph{Implication}
A key intuition is that if a relation exists,
then its two argument entities must also exist.
In other words,
it is impossible for a relation to exist without two corresponding entities.
From the perspective of probability,
it implies that the probability of relation is not greater than the probability of each argument entity.
Since we model entity and relation labels in a unified probability space,
this idea can be easily used in our model as the implication constraint.
We impose this constraint on $\mathcal{P}$:
for each word in the diagonal,
its maximum possibility over the entity type space $\mathcal{Y}_e$ must not be lower than the maximum possibility
for other words in the same row or column over the relation type space $\mathcal{Y}_r$.
We formulate this table-level constraint as implication loss,
\begin{IEEEeqnarray*}{rccl}
    \mathcal{L}_\mathrm{imp} & = & \frac{1}{|s|} & \sum_{i = 1}^{|s|}
    \left [\max_{l \in \mathcal{Y}_r} \{\mathcal{P}_{i,:,l}, \mathcal{P}_{:, i, l}\} - \max_{t \in \mathcal{Y}_e} \{\mathcal{P}_{i,i,t}\}\right]_*
\end{IEEEeqnarray*}
where $[u]_* = \max(u, 0)$ is the hinge loss.
It is worth noting that we do not add margin in this loss function.
Since the value of each item is a probability and might be relatively small,
it is meaningless to set a large margin.

Finally, we jointly optimize the three objectives in the training stage as $\mathcal{L}_\mathrm{entry} + \mathcal{L}_\mathrm{sym} + \mathcal{L}_\mathrm{imp}$.\footnote{We directly sum the three losses to avoid introducing more hyper-parameters.}


\subsection{Decoding} \label{joint-decoding}
In the testing stage,
given the probability tensor $\mathcal{P} \in \mathbb{R}^{|s| \times |s| \times |\mathcal{Y}|}$ of the sentence $s$,
\footnote{For the symmetrical label $t \in \mathcal{Y}_\mathrm{sym}$, we set $
        \mathcal{P}_{i,j,t} = \mathcal{P}_{j,i,t}  =  (\mathcal{P}_{i,j,t} + \mathcal{P}_{j,i,t}) / 2$.}
how to decode all rectangles (including squares) corresponding to entities or relations remains a non-trivial problem.
Since brute force enumeration of all rectangles is intractable,
a new joint decoding algorithm is needed.
We expect our decoder to have,
\begin{itemize}[leftmargin=1pc, itemindent=1pc]
    \item Simple implementation and fast decoding. We permit slight decoding accuracy drops for scalability.
    \item Strong interactions between entities and relations.
          When decoding entities, it should take the relation information into account, and vice versa.
\end{itemize}

\begin{figure}[t]
    \begin{center}
        \includegraphics[width=.98\columnwidth]{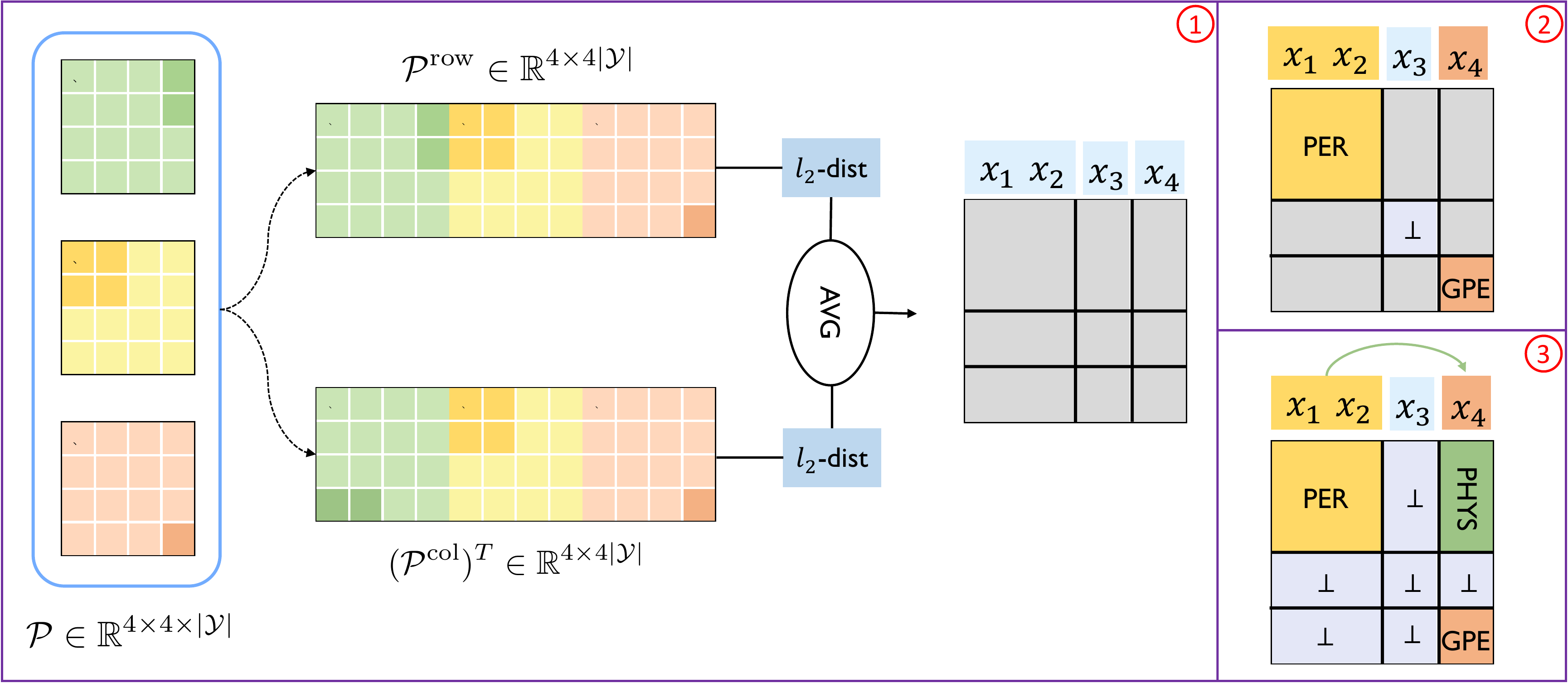}
    \end{center}
    \caption{Overview of our joint decoding algorithm. It consists of three steps: span decoding, entity type decoding, and relation type decoding.}
    \label{fig:decoding}
\end{figure}

Inspired by the procedures of \citep{sun-etal-2019-joint},
We propose a three-steps decoding algorithm:
decode \emph{span} first (entity spans or spans between entities),
and then decode \emph{entity type} of each span,
and at last decode \emph{relation type} of each entity pair
(\autoref{fig:decoding}).
We consider each cell's probability scores 
on all labels (including entity labels and relation labels) 
and predict spans according to a threshold.
Then, we predict entities and relations with the highest score.
Our heuristic decoding algorithm could be very efficient.
Next we will detail the entire decoding process,
and give a formal description in the \autoref{appendix:decoding-alg}.

\paragraph{Span Decoding}
One crucial observation of a ground-truth table is that,
for an arbitrary entity, its corresponding rows (or columns)
are exactly the same in the table
(e.g., row 1 and row 2 of \autoref{fig:task} are identical),
not only for the diagonal entries (entities are squares),
but also for the off-diagonal entries
(if it participates in a relation with another entity, all its rows (columns) will
spot that relation label in the same way).
In other words, if the adjacent rows/columns are different,
there must be an entity boundary
(i.e., one belonging to the entity and the other not belonging to the entity).
Therefore, if our biaffine model is reasonably trained,
given a model predicted table,
we could use this property to find split positions
of entity boundary.
As expected, experiments (Figure~\ref{fig:dist-distribution-ace05}) verify our assumption.
We adapt this idea to the 3-dimensional probability tensor $\mathcal{P}$.


Specifically,
we flatten $\mathcal{P} \in \mathbb{R}^{|s| \times |s| \times |\mathcal{Y}|}$ as a matrix $\mathcal{P}^\mathrm{row} \in \mathbb{R}^{|s| \times (|s| \cdot|\mathcal{Y}|)}$ from row perspective,
and then calculate the Euclidean distances ($l_2$ distances) of adjacent rows.
Similarly, we calculate the other Euclidean distances of adjacent columns according to a matrix $\mathcal{P}^\mathrm{col} \in
    \mathbb{R}^{(|s| \cdot|\mathcal{Y}|) \times |s|}$ from column perspective,
and then average the two distances as the final distance.
If the distance is larger than the threshold $\alpha$ ($\alpha = 1.4$ in our default settings),
this position is a split position.
In this way, we can decode all the spans in $\mathcal{O}(|s|)$ time complexity.

\paragraph{Entity Type Decoding}
Given a span $(i, j)$ by span decoding,\footnote{$i$ and $j$ denote start and end indices of the span.}
we decode the entity type $\hat{t}$ according to the corresponding square symmetric about the diagonal:
$\hat{t} = {\arg \max}_{t \in \mathcal{Y}_e \cup \{\bot\}}
    \mathtt{Avg}(\mathcal{P}_{i:j, i:j, t})$.
If $\hat{t} \in \mathcal{Y}_e$, we decode an entity.
If $\hat{t} = \bot$, the span $(i, j)$ is not an entity.

\paragraph{Relation Type Decoding}
After entity type decoding,
given an entity $e_1$ with the span $(i, j)$
and another entity $e_2$ with the span $(m, n)$,
we decode the relation type $\hat{l}$ between $e_1$ and $e_2$
according to the corresponding rectangle.
Formally,
$\hat{l} = {\arg \max}_{l \in \mathcal{Y}_r \cup \{\bot\}}
    \mathtt{Avg}(\mathcal{P}_{i:j, m:n, l})$.
If $\hat{l} \in \mathcal{Y}_r$, we decode a relation $(e_1, e_2, \hat{l})$.
If $\hat{l} = \bot$, $e_1$ and $e_2$ have no relation.

\section{Experiments}
\begin{table}[tb!]
    \centering
    \footnotesize
    \begin{tabular}{cccc}
        \toprule
        Dataset      & \#sents & \#ents(\#types) & \#rels(\#types) \\
        \midrule
        ACE04  & 8,683   & 22,519(7)       & 4,417(6)        \\
        ACE05  & 14,525  & 38,287(7)       & 7,691(6)        \\
        SciERC & 2,687   & 8,094(6)        & 5,463(7)        \\
        \bottomrule
    \end{tabular}
    \caption{The statistics of the adopted datasets.}
    \label{tab:data-statistics}
\end{table}
\begin{table*}[tb!]
    \centering
    \footnotesize
    \begin{tabular}{cllcccccc}
        \toprule
        \multirow{2}{*}{\textbf{Dataset}} & \multirow{2}{*}{\textbf{Model}}                            & \multirow{2}{*}{\textbf{Encoder}} & \multicolumn{3}{c}{\textbf{Entity}} & \multicolumn{3}{c}{\textbf{Relation}}                                                                 \\
        ~                                 & ~                                                          & ~                                 & P                                   & R                                     & F1            & P             & R             & F1            \\
        \midrule
        \multirow{9}{*}{\textbf{ACE04}}   & \citet{li-ji-2014-incremental}                             & -                                 & 83.5                                & 76.2                                  & 79.7          & 60.8          & 36.1          & 45.3          \\
        ~                                 & \citet{miwa-bansal-2016-end}                               & LSTM                              & 80.8                                & 82.9                                  & 81.8          & 48.7          & 48.1          & 48.4          \\
        ~                                 & \citet{katiyar-cardie-2017-going}                          & LSTM                              & 81.2                                & 78.1                                  & 79.6          & 46.4          & 45.3          & 45.7          \\
        ~                                 & \citet{li-etal-2019-entity}                                & BERT$_{\texttt{LARGE}}$           & 84.4                                & 82.9                                  & 83.6          & 50.1          & 48.7          & 49.4          \\
        ~                                 & \citet{wang-lu-2020-two}                                   & ALBERT$_{\texttt{XXLARGE}}$       & -                                   & -                                     & 88.6          & -             & -             & 59.6          \\
        ~                                 & \citet{zhong2020frustratingly}\textsuperscript{$\diamond$} & BERT$_{\texttt{BASE}}$            & -                                   & -                                     & 89.2          & -             & -             & 60.1          \\
        ~                                 & \citet{zhong2020frustratingly}\textsuperscript{$\diamond$} & ALBERT$_{\texttt{XXLARGE}}$       & -                                   & -                                     & \textbf{90.3} & -             & -             & 62.2          \\
        \cmidrule(lr){2-9}
        ~                                 & \mymodel\textsuperscript{$\diamond$}                           & BERT$_{\texttt{BASE}}$            & 87.4                                & 88.0                                  & 87.7          & 62.1          & 58.0          & 60.0          \\
        ~                                 & \mymodel\textsuperscript{$\diamond$}                           & ALBERT$_{\texttt{XXLARGE}}$       & \textbf{88.9}                       & \textbf{90.0}                         & 89.5          & \textbf{67.3} & \textbf{59.3} & \textbf{63.0} \\
        \midrule
        \multirow{11}{*}{\textbf{ACE05}}  & \citet{li-ji-2014-incremental}                             & -                                 & 85.2                                & 76.9                                  & 80.8          & 65.4          & 39.8          & 49.5          \\
        ~                                 & \citet{miwa-bansal-2016-end}                               & LSTM                              & 82.9                                & 83.9                                  & 83.4          & 57.2          & 54.0          & 55.6          \\
        ~                                 & \citet{katiyar-cardie-2017-going}                          & LSTM                              & 84.0                                & 81.3                                  & 82.6          & 55.5          & 51.8          & 53.6          \\
        ~                                 & \citet{sun-etal-2019-joint}                                & LSTM                              & 86.1                                & 82.4                                  & 84.2          & 68.1          & 52.3          & 59.1          \\
        ~                                 & \citet{li-etal-2019-entity}                                & BERT$_{\texttt{LARGE}}$           & 84.7                                & 84.9                                  & 84.8          & 64.8          & 56.2          & 60.2          \\
        ~                                 & \citet{wang-etal-2020-pre}                                 & BERT$_{\texttt{BASE}}$            & -                                   & -                                     & 87.2          & -             & -             & 63.2          \\
        ~                                 & \citet{wang-lu-2020-two}                                   & ALBERT$_{\texttt{XXLARGE}}$       & -                                   & -                                     & 89.5          & -             & -             & 64.3          \\
        ~                                 & \citet{zhong2020frustratingly}\textsuperscript{$\diamond$} & BERT$_{\texttt{BASE}}$            & -                                   & -                                     & 90.2          & -             & -             & 64.6          \\
        ~                                 & \citet{zhong2020frustratingly}\textsuperscript{$\diamond$} & ALBERT$_{\texttt{XXLARGE}}$       & -                                   & -                                     & \textbf{90.9} & -             & -             & \textbf{67.8} \\
        \cmidrule(lr){2-9}
        ~                                 & \mymodel\textsuperscript{$\diamond$}                           & BERT$_{\texttt{BASE}}$            & 88.8                                & 88.9                                  & 88.8          & 67.1          & \textbf{61.8} & 64.3          \\
        ~                                 & \mymodel\textsuperscript{$\diamond$}                           & ALBERT$_{\texttt{XXLARGE}}$       & \textbf{89.9}                       & \textbf{90.5}                         & 90.2          & \textbf{72.3} & 60.7          & 66.0          \\
        \midrule
        \multirow{3}{*}{\textbf{SciERC}}
                                          & \citet{wang-etal-2020-pre}                                 & SciBERT                           & -                                   & -                                     & 68.0          & -             & -             & 34.6          \\
        ~                                 & \citet{zhong2020frustratingly}\textsuperscript{$\diamond$} &
        SciBERT                           & -                                                          & -                                 & 68.2                                & -                                     & -             & 36.7                                          \\
        \cmidrule(lr){2-9}
        ~                                 & \mymodel\textsuperscript{$\diamond$}                           & SciBERT                           & \textbf{65.8}                       & \textbf{71.1}                         & \textbf{68.4} & \textbf{37.3} & \textbf{36.6} & \textbf{36.9} \\
        \bottomrule
    \end{tabular}
    \caption{Overall evaluation. $\diamond$ means that the model leverages cross-sentence context information.}
    \label{tab:main-results}
\end{table*}

\paragraph{Datasets}
We conduct experiments on three entity relation extraction benchmarks:
ACE04 \citep{doddington-etal-2004-automatic},\footnote{\url{https://catalog.ldc.upenn.edu/LDC2005T09}}
ACE05 \citep{walker2006ace},\footnote{\url{https://catalog.ldc.upenn.edu/LDC2006T06}}
and SciERC \citep{luan-etal-2018-multi}.\footnote{\url{http://nlp.cs.washington.edu/sciIE/}}
\autoref{tab:data-statistics} shows the dataset statistics.
Besides, we provide detailed dataset specifications in the \autoref{appendix:datasets}.




\paragraph{Evaluation}
Following suggestions in \cite{taille-etal-2020-lets},
we evaluate Precision (P), Recall (R), and F1 scores with micro-averaging and adopt the \textbf{Strict Evaluation} criterion.
Specifically,
a predicted entity is correct if its type and boundaries are correct,
and a predicted relation is correct if its relation type is correct, as well as the boundaries and types of two argument entities are correct.

\paragraph{Implementation Details}
We tune all hyper-parameters based on the averaged entity F1 and relation F1 on ACE05 development set,
then keep the same settings on ACE04 and SciERC.
For fair comparison with previous works,
we use three pre-trained language models: \texttt{bert-base-uncased} \citep{devlin-etal-2019-bert}, \texttt{albert-xxlarge-v1} \citep{lan2019albert} and \texttt{scibert-scivocab-uncased} \citep{beltagy2019scibert} as the sentence encoder and fine-tune them in training stage.\footnote{The first two are for ACE04 and ACE05, and the last one is for SciERC.}

For the MLP layer,
we set the hidden size as $d = 150$ and use GELU as the activation function.
We use AdamW optimizer \citep{loshchilov2017decoupled} with $\beta_1 = 0.9$ and $\beta_2 = 0.9$,
and observe a phenomenon similar to \citep{dozat2016deep} in that setting $\beta_2$ from 0.9 to 0.999 causes a significant drop on final performance.
The batch size is 32, and the learning rate is 5e-5 with weight decay 1e-5.
We apply a linear warm-up learning rate scheduler with a warm-up ratio of 0.2.
We train our model with a maximum of 200 epochs (300 epochs for SciERC) and employ an early stop strategy.
We perform all experiments on an Intel(R) Xeon(R) W-3175X CPU and a NVIDIA Quadro RTX 8000 GPU.

\subsection{Performance Comparison}
\autoref{tab:main-results} summarizes previous works and our \mymodel{} on three datasets.\footnote{Since \cite{luan2019general, wadden-etal-2019-entity} neglect the argument entity type in relation evaluation and underperform our baseline \citep{zhang2020minimize}, we do not compare their results here.}
In general, \mymodel{} achieves the best performance on ACE04 and SciERC and a comparable result on ACE05.
Comparing with the previous best joint model \citep{wang-lu-2020-two},
our model significantly advances both entity and relation performances,
i.e., an absolute F1 of +0.9 and +0.7 for entity as well as +3.4 and +1.7 for relation,
on ACE04 and ACE05 respectively.
For the best pipeline model \citep{zhong2020frustratingly} (current SOTA),
our model achieves superior performance on ACE04 and SciERC and comparable performance on ACE05.
Comparing with ACE04/ACE05, SciERC is much smaller, so entity performance on SciERC drops sharply. 
Since \citep{zhong2020frustratingly} is a pipeline method, 
its relation performance is severely influenced by the poor entity performance. 
Nevertheless, our model is less influenced in this case and achieves better performance.
Besides, our model can achieve better relation performance even with worse entity results on ACE04.
Actually, our base model (BERT$_{\texttt{BASE}}$) has achieved competitive relation performance,
which even exceeds prior models based on BERT$_{\texttt{LARGE}}$ \cite{li-etal-2019-entity} and ALBERT$_{\texttt{XXLARGE}}$ \cite{wang-lu-2020-two}.
These results confirm the proposed unified label space is effective for exploring the interaction between entities and relations.
Note that all subsequent experiment results on ACE04 and ACE05 are based on BERT$_{\texttt{BASE}}$ for efficiency.

\begin{table}[t]
    \centering
    \footnotesize
    \begin{tabular}{lcccc}
        \toprule
        \multirow{2}{*}{\textbf{Settings}} & \multicolumn{2}{c}{\textbf{ACE05}} & \multicolumn{2}{c}{\textbf{SciERC}}                                 \\
        ~                                  & Ent                                & Rel                                 & Ent           & Rel           \\
        \midrule
        Default                            & 88.8                               & \textbf{64.3}                       & \textbf{68.4} & 36.9          \\
        \midrule
        w/o symmetry loss                  & 88.9                               & 64.0                                & 67.3          & 35.5          \\
        w/o implication loss               & \textbf{89.0}                      & 63.3                                & 68.0          & \textbf{37.1} \\
        w/o logit dropout                  & 88.8                               & 61.8                                & 66.9          & 34.7          \\
        w/o cross-sentence context         & 87.9                               & 62.7                                & 65.3          & 32.1          \\
        \midrule
        hard decoding                      & 74.0                               & 34.6                                & 46.1          & 17.8          \\
        \bottomrule
    \end{tabular}
    \caption{Results (F1 score) with different settings on ACE05 and SciERC test sets. Note that we use BERT$_{\texttt{BASE}}$ on ACE05.}
    \label{tab:ablation}
\end{table}

\subsection{Ablation Study}
In this section, we analyze the effects of components in \mymodel{} with different settings (\autoref{tab:ablation}).
Particularly, we implement a naive decoding algorithm for comparison,
namely ``hard decoding'',
which takes the ``intermediate table'' as input.
The ``intermediate table'' is the hard form of probability tensor $\mathcal{P}$ output by the biaffine model,
i.e., choosing the class with the highest probability as the label of each cell.
To find entity squares on the diagonal,
it first tries to judge whether the largest
square ($|s|\times|s|$) is an entity.
The criterion is simply counting the
number of different entity labels appearing in the square
and choosing the most frequent one.
If the most frequent label is $\bot$,
we shrink the size of square by $1$ and do the same work
on two $(|s|-1)\times(|s|-1)$ squares and so on.
To avoid entity overlapping, an entity will be discarded if it overlaps with identified entities.
To find relations,
each entity pair is labeled by the most frequent relation label in the
corresponding rectangle.

From the ablation study, we get the following observations.

\begin{table}[t]
    \centering
    \footnotesize
    \resizebox{0.48\textwidth}{!}{
        \begin{tabular}{lccccccc}
            \toprule
            \multirow{2}{*}{\textbf{Model}}                  & \multirow{2}{*}{\textbf{Parameters}} & \multirow{2}{*}{\textbf{W}} & \multicolumn{2}{c}{\textbf{ACE05}} & \multicolumn{2}{c}{\textbf{SciERC}}                                  \\
            ~                                                & ~                                    & ~                           & \makecell[c]{Rel                                                                                          \\(F1)}                                                         & \makecell[c]{Speed\\(sent/s)} &  \makecell[c]{Rel\\(F1)} & \makecell[c]{Speed\\(sent/s)} \\
            \midrule
            Z\&C(\citeyear{zhong2020frustratingly})          & 219M                                 & 100                         & \textbf{64.6}                      & 14.7                                & 36.7          & 19.9           \\
            Z\&C(\citeyear{zhong2020frustratingly})$\dagger$ & 219M                                 & 100                         & -                                  & 237.6                               & -             & 194.7          \\
            \midrule
            \mymodel{}                                       & 110M                                 & 100                         & 63.6                               & \textbf{340.6}                      & 34.0          & \textbf{314.8} \\
            \mymodel{}                                       & 110M                                 & 200                         & 64.3                               & 194.2                               & \textbf{36.9} & 200.1          \\
            \midrule
            hard decoding                                    & 110M                                 & 200                         & 34.6                               & 139.1                               & 17.8          & 113.0          \\
            \bottomrule
        \end{tabular}}
    \caption{Comparison of accuracy and efficiency on ACE05 and SciERC test sets with different context window sizes. $\dagger$ denotes the approximation version with a faster speed and a worse performance.}
    \label{tab:speed}
\end{table}

\begin{itemize}[leftmargin=1pc, itemindent=1pc]
    \item When one of the additional losses is removed, the performance will decline with varying degrees (line 2-3). Specifically, the symmetrical loss has a significant impact on SciERC (decrease 1.1 points and 1.4 points for entity and relation performance). While removing the implication loss will obviously harm the relation performance on ACE05 (1.0 point). It demonstrates that the structural information incorporated by both losses is useful for this task.
    \item Comparing with the ``Default'', the performance of  ``w/o logit dropout'' and ``w/o cross-sentence context'' drop more sharply (line 4-5). Logit dropout prevents the model from overfitting, and cross-sentence context provides more contextual information for this task, especially for small datasets like SciERC.
    \item The ``hard decoding'' has the worst performance (its relation performance is almost half of the ``Default'') (line 6). The major reason is that ``hard decoding'' separately decodes entities and relations. It shows the proposed decoding algorithm jointly considers entities and relations, which is important for decoding.
\end{itemize}

\subsection{Inference Speed}
Following \citep{zhong2020frustratingly},
we evaluate the inference speed of our model (\autoref{tab:speed}) on ACE05 and SciERC with the same batch size and pre-trained encoders (BERT$_{\texttt{BASE}}$ for ACE05 and SciBERT for SciERC).
Comparing with the pipeline method \citep{zhong2020frustratingly},
we obtain a more than $10\times$ speedup and achieve a comparable or even better relation performance with $W = 200$.
As for their approximate version,
our inference speed is still competitive but with better performance.
If the context window size is set the same as \citep{zhong2020frustratingly} ($W=100$),
we can further accelerate model inference with slight performance drops.
Besides, ``hard decoding'' is much slower than \mymodel{},
which demonstrates the efficiency of the proposed decoding algorithm.

\subsection{Impact of Different Threshold $\alpha$}
\begin{figure}[t]
    \begin{center}
        \includegraphics[width=2.3in]{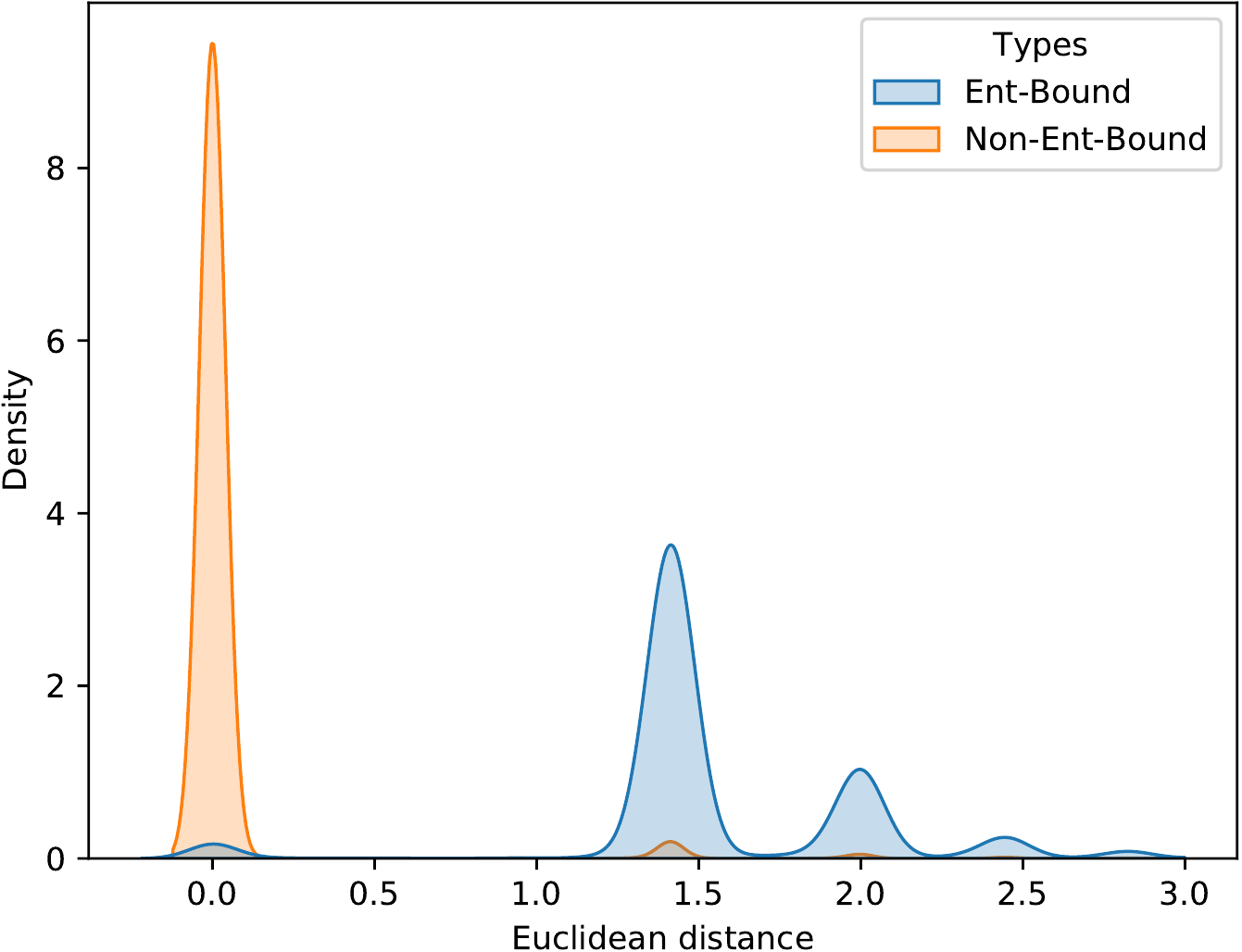}
    \end{center}
    \caption{Distributions of adjacent rows' distances for two categories with respect to the threshold $\alpha$ on ACE05 dev set.}
    \label{fig:dist-distribution-ace05}
\end{figure}

\begin{figure}[t]
    \begin{center}
        \includegraphics[width=2.3in]{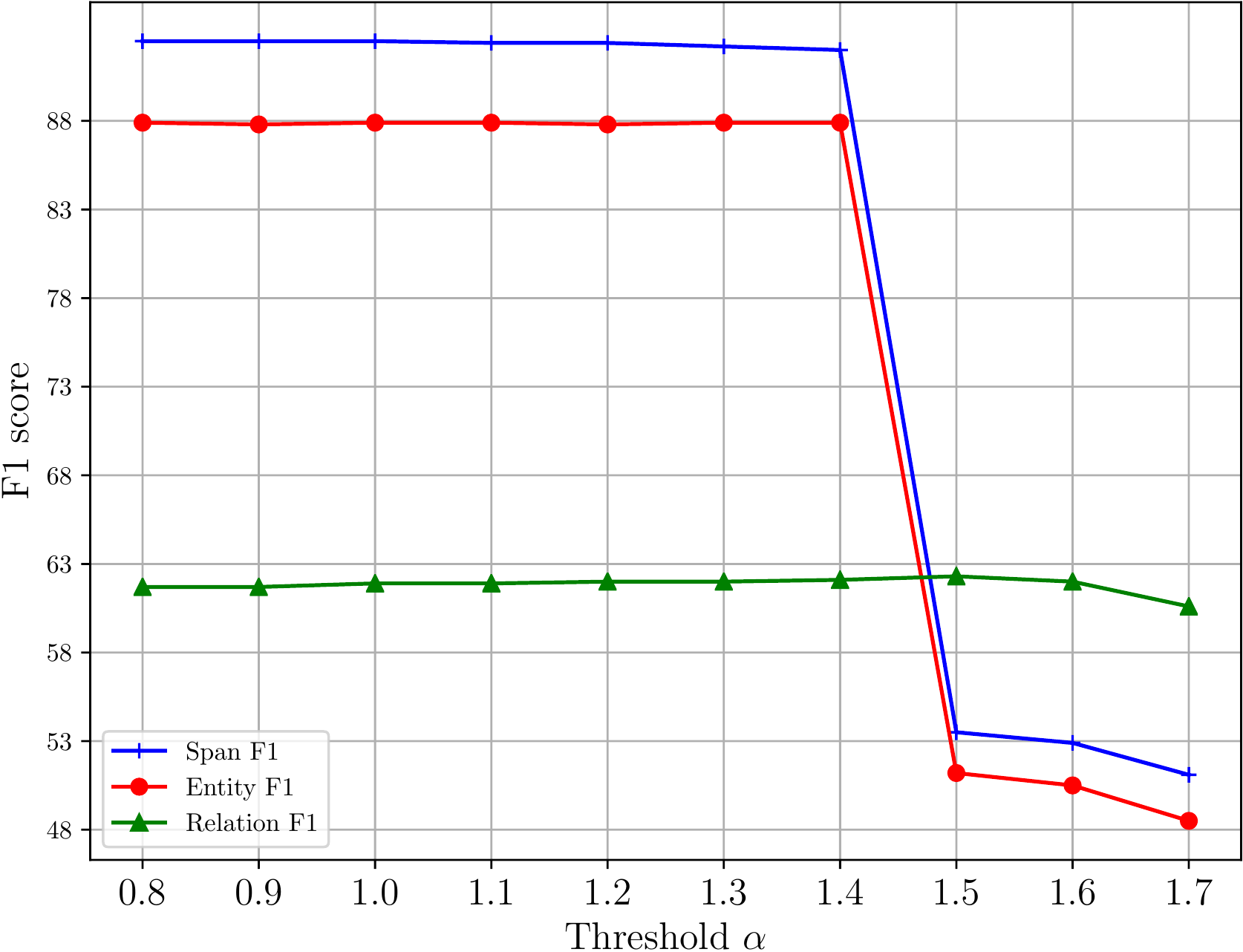}
    \end{center}
    \caption{Performances with respect to the threshold $\alpha$ on ACE05 dev set.}
    \label{fig:split-threshold-ace05}
\end{figure}

In Figure~\ref{fig:dist-distribution-ace05},
the distance between adjacent rows not at entity boundary (``Non-Ent-Bound'') mainly concentrates at 0,
while that at entity boundary (``Ent-Bound'') is usually greater than 1.
This phenomenon verifies the correctness of our span decoding method.
Then we evaluate the performances,
with regard to the threshold $\alpha$ in Figure~\ref{fig:split-threshold-ace05}.\footnote{We use an additional metric to evaluate span performance, ``Span F1'', is Micro-F1 of predicted split positions.} 
Both span and entity performances sharply decrease when $\alpha$ increases from 1.4 to 1.5,
while the relation performance starts to decline slowly from $\alpha=1.5$.
The major reason is that relations are so sparse that many entities do not participate in any relation,
so the threshold of relation is much higher than that of entity.
Moreover, we observe a similar phenomenon on ACE04 and SciERC, and $\alpha=1.4$  is a general best setting on three datasets.
It shows the stability and generalization of our model.


\subsection{Context Window and Logit Dropout Rate}
In \autoref{tab:ablation}, both cross-sentence context and logit dropout can improve the entity and relation performance.
\autoref{tab:hyper-parameters} shows the effect of different context window size $W$ and logit dropout rate $p$.
The entity and relation performances are significantly improved from $W = 100$ to $W = 200$,
and drop sharply from $W = 200$ to $W = 300$.
Similarly,
we achieve the best entity and relation performances when $p = 0.2$.
So we use $W = 200$ and $p = 0.2$ in our final model.

\begin{table}[tb!]
    \centering
    \footnotesize
    \resizebox{0.4\textwidth}{!}{\begin{tabular}{lccccc}
        \toprule
        ~                    & \multirow{2}{*}{\textbf{Value}} & \multicolumn{2}{c}{\textbf{ACE05}} & \multicolumn{2}{c}{\textbf{SciERC}}                                 \\
        ~                    & ~                               & Ent                                & Rel                                 & Ent           & Rel           \\
        \midrule
        \multirow{3}{*}{$W$} & 100                             & 87.4                               & \textbf{62.4}                       & 69.0          & 36.7          \\
        ~                    & 200                             & \textbf{87.9}                      & 62.1                                & \textbf{70.6} & \textbf{38.3} \\
        ~                    & 300                             & 87.2                               & 60.8                                & 69.4          & 35.4          \\
        \midrule
        \multirow{4}{*}{$p$} & 0.1                             & 87.4                               & 61.8                                & \textbf{71.1} & 37.8          \\
        ~                    & 0.2                             & \textbf{87.9}                      & \textbf{62.1}                       & 70.6          & \textbf{38.3} \\
        ~                    & 0.3                             & 87.2                               & \textbf{62.1}                       & 67.8          & 33.5          \\
        ~                    & 0.4                             & 87.4                               & 62.0                                & 70.6          & 35.8          \\
        \bottomrule
    \end{tabular}}
    \caption{Results (F1 scores) with respect to the context window size and the logit dropout rate on ACE05 and SciERC dev sets.}
    \label{tab:hyper-parameters}
\end{table}

\subsection{Error Analysis}
\begin{figure}[t]
    \begin{center}
        \includegraphics[width=2.5in]{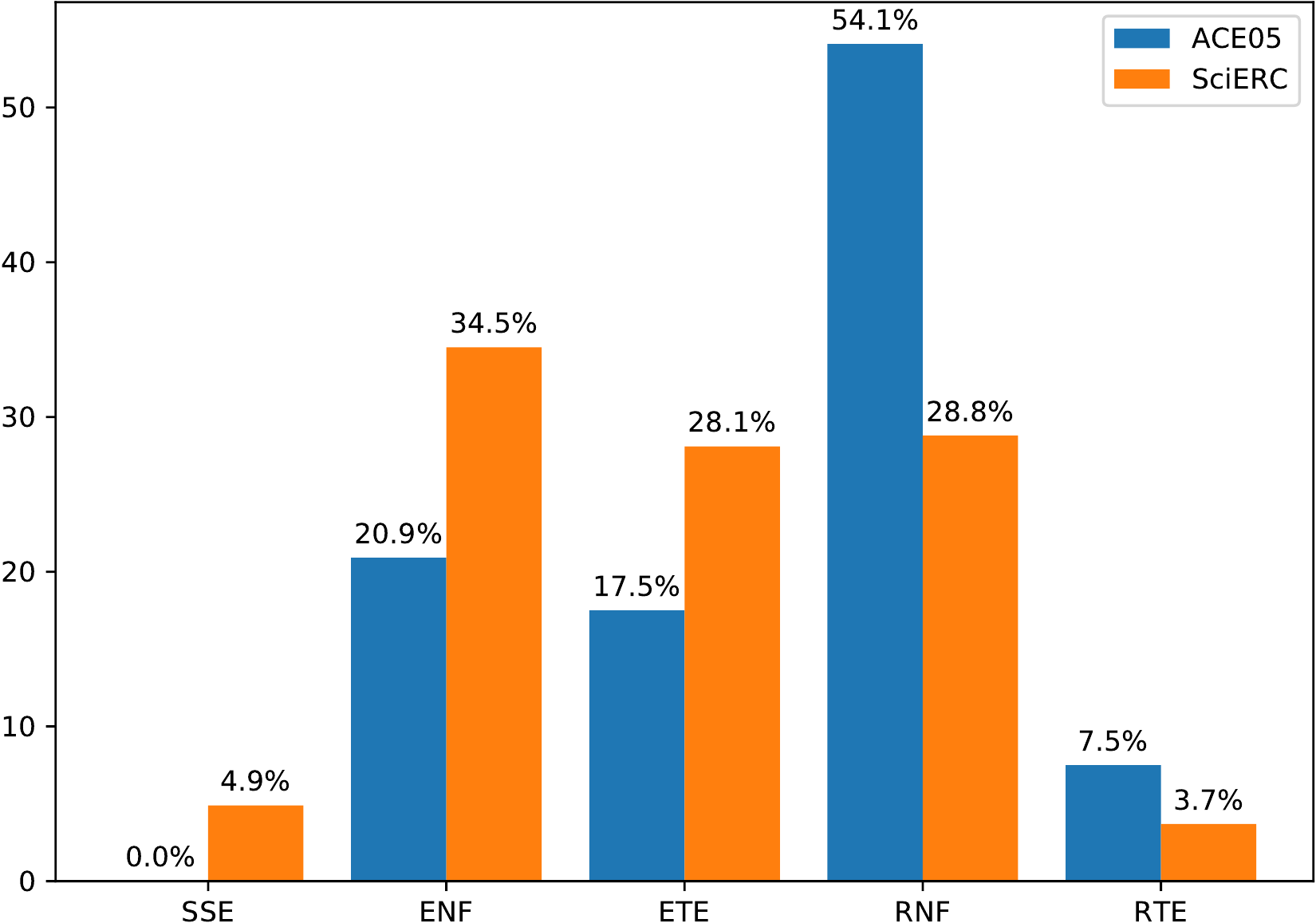}
    \end{center}
    \caption{Distribution of five relation extraction errors on ACE05 and SciERC test data.}
    \label{fig:error-analysis}
\end{figure}

We further analyze the remaining errors for relation extraction and present the distribution of five errors: span splitting error (SSE), entity not found (ENF), entity type error (ETE), relation not found (RNF), and relation type error (RTE) in \autoref{fig:error-analysis}.
The proportion of ``SSE'' is relatively small, which proves the effectiveness of our span decoding method.
Moreover,
the proportion of ``not found error'' is significantly larger than that of ``type error'' for both entity and relation.
The primary reason is that the table filling suffers from the class imbalance issue,
i.e.,
the number of $\bot$ is much larger than that of other classes.
We reserve this imbalanced classification problem in the future.

\begin{figure}[t]
    \begin{center}
        \includegraphics[width=2.5in]{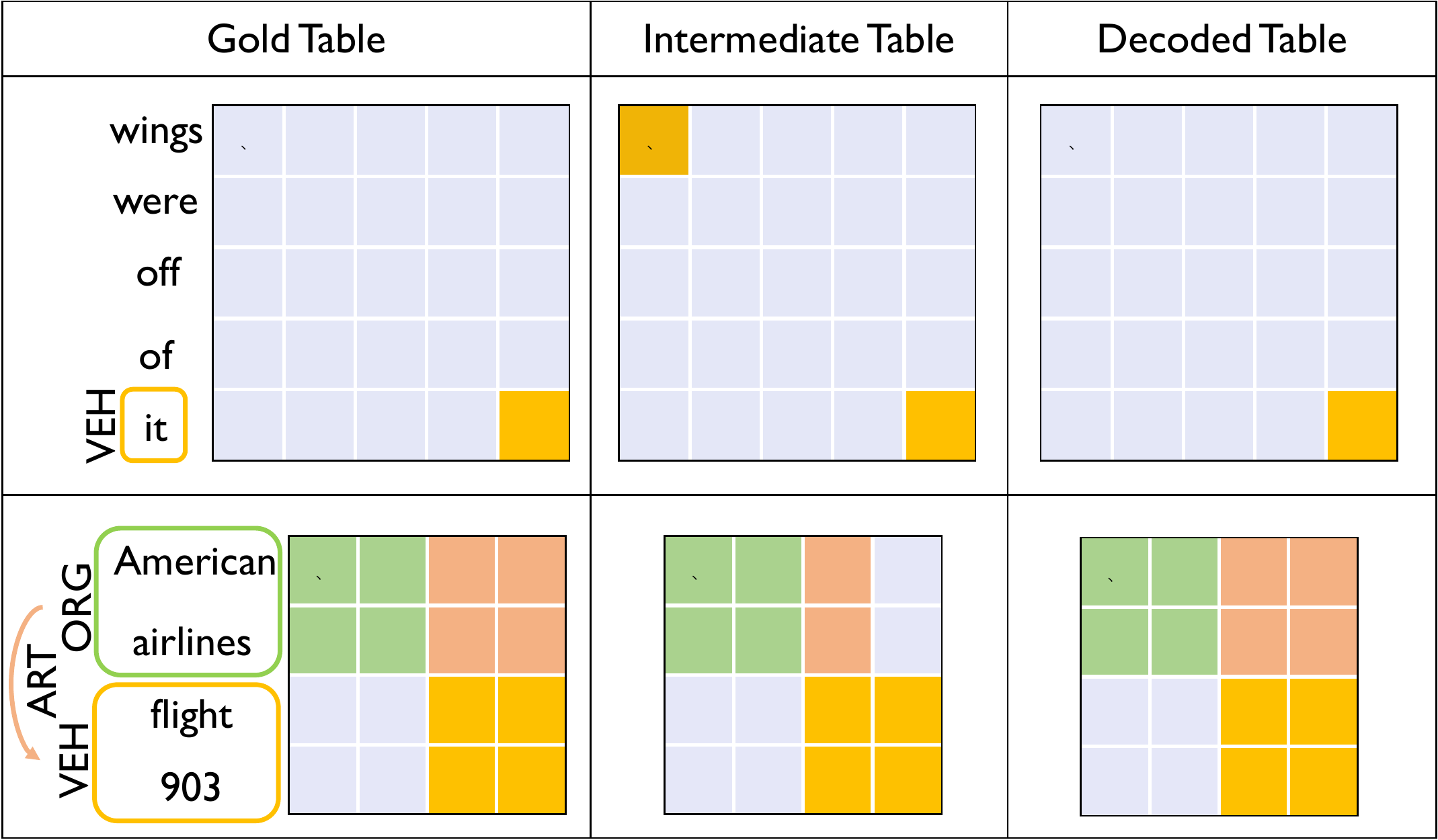}
    \end{center}
    \caption{Examples showing the robustness of our decoding algorithm. ``Gold Table'' presents the gold label. ``Intermediate Table'' presents the biaffine model's prediction (choosing the label with the highest probability for each cell). ``Decoded Table'' presents the final results after decoding.}
    \label{fig:decoding-case}
\end{figure}

Finally, we give some concrete examples in \autoref{fig:decoding-case} to verify the robustness of our decoding algorithm.
There are some errors in the biaffine model's prediction,
such as cells in the upper left corner (first example) and upper right corner (second example) in the intermediate table.
However, these errors are corrected after decoding,
which demonstrates that our decoding algorithm not only recover all entities and relations 
but also corrects errors leveraging table structure and neighbor cells' information.

\section{Related Work}
Entity relation extraction has been extensively studied over the decades.
Existing methods can be roughly divided into two categories according to the adopted label space.

\paragraph{Separate Label Spaces}
This category study this task as two separate sub-tasks:
entity recognition and relation classification,
which are defined in two separate label spaces.
One early paradigm is the pipeline method \citep{zelenko2003kernel,miwa-etal-2009-rich} that uses two independent models for two sub-tasks respectively.
Then joint method handles this task with an end-to-end model to explore more interaction between entities and relations.
The most basic joint paradigm,
parameter sharing \citep{miwa-bansal-2016-end, katiyar-cardie-2017-going},
adopts two independent decoders based on a shared encoder.
Recent span-based models \citep{luan-etal-2019-general, wadden-etal-2019-entity} also use this paradigm.
To enhance the connection of two decoders,
many joint decoding algorithms are proposed,
such as ILP-based joint decoder \citep{yang2013joint}, joint MRT \cite{sun-etal-2018-extracting}, GCN-based joint inference \citep{sun-etal-2019-joint}.
Actually, table filling method \citep{miwa2014modeling,gupta-etal-2016-table,zhang-etal-2017-end,wang-etal-2020-pre} is a special case of parameter sharing in table structure.
These joint models all focus on various joint algorithms but ignore the fact that they are essentially based on separate label spaces.

\paragraph{Unified Label Space}
This family of methods aims to unify two sub-tasks and tackle this task in a unified label space.
Entity relation extraction has been converted into a tagging problem \citep{zheng2017joint}, a transition-based parsing problem \citep{wang2018joint}, and a generation problem with Seq2Seq framework \citep{zeng2018extracting, nayak2020effective}.
We follow this trend and propose a new unified label space.
We introduce a 2D table to tackle the overlapping relation problem in \citep{zheng2017joint}.
Also, our model is more versatile as not relying on complex expertise like \citep{wang2018joint},
which requires external expert knowledge to design a complex transition system.


\section{Conclusion}
In this work, we extract entities and relations in a unified label space to better mine
the interaction between both sub-tasks.
We propose a novel table that presents entities and relations as squares and rectangles.
Then this task can be performed in two simple steps:
filling the table with our biaffine model and decoding entities and relations with our joint decoding algorithm.
Experiments on three benchmarks show the proposed method achieves not only state-of-the-art performance but also promising efficiency.

\section*{Acknowledgement}
The authors wish to thank the reviewers for their helpful comments and suggestions.
This work was (partially) supported by National Key Research and Development Program of China (2018AAA0100704), NSFC (61972250, 62076097), STCSM (18ZR1411500), Shanghai Municipal Science and Technology Major Project (2021SHZDZX0102), and the Fundamental Research Funds for the Central Universities.

\bibliography{joint-decoding-entrel}
\bibliographystyle{acl_natbib}

\newpage
\appendix
\section{Decoding Algorithm}
\label{appendix:decoding-alg}
A formal description are shown in Algorithm~\autoref{alg:joint-decoding-algorithm}.

\begin{algorithm}[t]
    \caption{Decoding Algorithm}
    \small
    \label{alg:joint-decoding-algorithm}
    \begin{algorithmic}[1]
        \Require
        Probability tensor $\mathcal{P} \in \mathbb{R}^{|s| \times |s| \times |\mathcal{Y}|}$ of sentence $s$
        \Ensure
        A set of entities $\mathcal{E}$ and a set of relations $\mathcal{R}$
        \State $\mathcal{E}_\mathrm{split} = []$, $\mathcal{E} = \mathtt{set}()$, $\mathcal{R} = \mathtt{set}()$
        \State $\mathcal{P}^\mathrm{row} \leftarrow \mathcal{P}.\mathtt{view}(n, n * |\mathcal{Y}|)$
        \State $\mathcal{P}^\mathrm{col} \leftarrow \mathcal{P}.\mathtt{transpose}(0, 1).\mathtt{view}(n, n * |\mathcal{Y}|)$
        \State $i \leftarrow 1$
        \While {$i \textless |s|$}
        \State $d \leftarrow \frac{||\mathcal{P}^\mathrm{row}_i - \mathcal{P}^\mathrm{row}_{i+1}||_2^2 + ||\mathcal{P}^\mathrm{col}_i - \mathcal{P}^\mathrm{col}_{i+1}||_2^2}{2}$
        \If {$ d \textgreater \alpha$}
        \State $\mathcal{E}_\mathrm{split}.\mathtt{append}(i)$
        \EndIf
        \State $i \leftarrow i + 1$
        \EndWhile
        \State $\mathcal{E}_\mathrm{split}.\mathtt{append}(|s|)$
        \State $i \leftarrow 1$
        \For {$j \in \mathcal{E}_\mathrm{split}$}
        \State $\hat{t} = {\arg \max}_{t \in \mathcal{Y}_e \cup \{\bot\}}\mathtt{Avg}(\mathcal{P}_{i:j, i: j, t})$
        \If {$\hat{t} \neq \bot$}
        \State $\mathtt{new}\ e$:  $e.\mathtt{span} = (i, j) $ and $e.\mathtt{type} = \hat{t}$
        \State $\mathcal{E}.\mathtt{add}(e)$
        \EndIf
        \State $i \leftarrow j + 1$
        \EndFor
        \For {$e_1, e_2 \in \mathcal{E}, e_1 \neq e_2$}
        \State $(i, j) = e_1.\mathtt{span}$
        \State $(m, n) = e_2.\mathtt{span}$
        \State $\hat{l} = {\arg \max}_{l \in \mathcal{Y}_r \cup \{\bot\}}\mathtt{Avg}(\mathcal{P}_{i:j, m:n, l})$
        \If {$\hat{l} \neq \bot$}
        \State $\mathcal{R}.\mathtt{add}((e_1, e_2, \hat{l}))$
        \EndIf
        \EndFor
    \end{algorithmic}
\end{algorithm}

\section{Datasets}
\label{appendix:datasets}
The ACE04 and ACE05 corpora are collected from various domains, such as newswire and online forums.
Both corpora annotate 7 entity types and 6 relation types.
we use the same data splits and pre-processing as \citep{li-ji-2014-incremental, miwa-bansal-2016-end},
i.e., 5-fold cross-validation for ACE04,
and 351 training, 80 validating, and 80 testing for ACE05.\footnote{We use the pre-processing scripts provided by \citep{wang-lu-2020-two} at \url{https://github.com/LorrinWWW/two-are-better-than-one/tree/master/datasets}.}
Besides, we randomly sample 10\% of training set as the development set for ACE04.

The SciERC corpus collects 500 scientiﬁc abstracts taken from AI conference/workshop proceedings.
This dataset annotates 6 entity types and 7 relation types.
We adopt the same data split protocol as in \citep{luan-etal-2019-general} (350 training, 50 validating, and 100 testing).
Detailed dataset specifications are shown in \autoref{tab:data-statistics}.

Moreover, we correct the annotations of undirected relations for three datasets,
regarding each undirected relation as two directed relation instances,
e.g., for the undirected relation \texttt{PER-SOC},
only one relation triplet (``his'', wife'', \texttt{PER-SOC}) is annotated in the original dataset,
we will add another relation triplet (``wife'', ``his'', \texttt{PER-SOC}) in our corrected datasets for symmetry.
In this case, each undirected relation corresponds to two rectangles, which are symmetrical about the diagonal.

\end{document}